\documentclass[journal]{IEEEtran}
\ifCLASSINFOpdf
\else
\fi
\usepackage{amssymb}
\usepackage{graphicx} 
\usepackage{amsthm}

\usepackage{comment} 
\usepackage{setspace} 
\usepackage{fullpage, epsfig, subfigure, wrapfig}
\usepackage{amssymb,amsmath}
\usepackage{tikz}

\newtheorem{definition}{Definition}

\usepackage[numbers]{natbib}
\usetikzlibrary{positioning}

\begin{document}
%

\title{\LARGE\bf LLM-Stackelberg Games: Conjectural Reasoning Equilibria and Their Applications to Spearphishing}
%
%
%

\author{Quanyan Zhu%
\thanks{Q. Zhu is with the Department of Electrical and Computer Engineering, New York University, Brooklyn, NY 11201 USA (e-mail: qz494@nyu.edu). This work was partially supported by NSF EPCN Award \#1847056.}%
}

\maketitle

\begin{abstract}
We introduce the framework of \emph{LLM-Stackelberg games}, a class of sequential decision-making models that integrate large language models (LLMs) into strategic interactions between a leader and a follower. Departing from classical Stackelberg assumptions of complete information and rational agents, our formulation allows each agent to reason through structured prompts, generate probabilistic behaviors via LLMs, and adapt their strategies through internal cognition and belief updates. We define two equilibrium concepts: \emph{reasoning and behavioral equilibrium}, which aligns an agent’s internal prompt-based reasoning with observable behavior, and \emph{conjectural reasoning equilibrium}, which accounts for epistemic uncertainty through parameterized  models over an opponent’s response. These layered constructs capture bounded rationality, asymmetric information, and meta-cognitive adaptation. We illustrate the framework through a spearphishing case study, where a sender and a recipient engage in a deception game using structured reasoning prompts. This example highlights the cognitive richness and adversarial potential of LLM-mediated interactions. Our results show that LLM-Stackelberg games provide a powerful paradigm for modeling decision-making in domains such as cybersecurity, misinformation, and recommendation systems.
\end{abstract}


%
\IEEEpeerreviewmaketitle

\section{Introduction}

Stackelberg games model sequential interactions between a leader and a follower, where the leader first commits to an action, and the follower subsequently responds optimally. This framework is foundational across domains such as cybersecurity \cite{kar2016trends,yang2025herd}, incentive design \cite{zhu2025revisiting,huang2020dynamic}, and adversarial prompt engineering for large language models (LLMs) \cite{zhu2025stackelberg}. In adversarial LLM scenarios, such as jailbreak attacks or manipulation of safety filters, the user serves as a strategic leader who creates input prompts, while the LLM responds in a restricted manner based on its training data, internal parameters, and alignment objectives.

LLMs enable the modeling of strategic interactions expressed in natural language, where both the actions and the responses are textual in nature. Their generative capacity supports simulation and optimization of complex messaging strategies, while their internal reasoning capabilities, often scaffolded through prompt templates such as chain-of-thought, make it possible to capture bounded rationality and structured cognition. This turns abstract assumptions about beliefs, preferences, and heuristics into tangible, configurable prompt-based reasoning procedures.

Classical Stackelberg models typically rely on strong assumptions: the leader is fully informed about the follower's behavior, utility functions are explicitly known, and agents are fully rational. However, such assumptions are often violated in real-world environments. Leaders rarely possess complete information about a follower's private context, internal reasoning model, or action policy. LLMs offer a new paradigm for reasoning under epistemic uncertainty, where agents form and refine conjectures about others' strategies within the reasoning process itself. These conjectures, expressed at either the behavioral or reasoning level, enable agents to simulate responses and adapt over time, thus supporting epistemic reasoning that extends beyond traditional aleatoric models of stochastic uncertainty \cite{li2025computational}.

We introduce LLM-Stackelberg games in which each agent selects a structured reasoning prompt that guides its interaction with an LLM. The leader (or sender) constructs a prompt to generate a message, while the follower (or receiver) processes that message with its own reasoning prompt to produce a response. These prompts serve as cognitive substrates that encode internal beliefs, heuristics, and processing styles, including inattention, bounded memory, or biases, especially when human cognitive patterns are involved. The resulting behavior is thus prompt-induced, and governed by the interplay between the LLM’s internal distribution and the reasoning structure encoded in the prompt.

The central equilibrium concept, rationalization and behavioral equilibrium, captures the consistency between the agent's internal mindset (including prior knowledge, reasoning strategy, and LLM worldview) \cite{zhu2025llmnash} and its observable behavior. At equilibrium, each agent optimally selects a reasoning prompt given its information and expectations, and the resulting message-action pair reflects a fixed point in this joint reasoning space. When agents lack complete knowledge of the internal models of each other, conjectural Stackelberg equilibria emerge: The leader forms a parameterized belief in the follower's response model and optimizes under this belief, refining it through observed outcomes and minimizing divergence from actual behavior. This LLM-augmented Stackelberg formulation naturally generalizes to multi-agent cognition, where each agent reasons not only about the environment but also about other agents' reasoning strategies. It also supports meta-cognition \cite{metcalfe1994metacognition}, where an agent reasons about its own cognitive processes, enabling self-evaluation, feedback-driven adaptation, and recursive improvement over time. 

We illustrate the framework through a spearphishing case study, in which an attacker LLM uses structured prompts to generate messages targeting a specific recipient. The recipient, modeled as another LLM-powered agent, evaluates the message using a diagnostic reasoning prompt and responds based on contextual cues, perceived risk, and cognitive heuristics. This example demonstrates the capability of LLM-Stackelberg games to model spearphishing interactions and generate strategically deceptive messages designed to manipulate recipients. It highlights the need for effective countermeasures and defense mechanisms.

Section II reviews classical Stackelberg game theory and formalizes the rational response set and Stackelberg equilibrium. Section III introduces the formulation of the LLM-Stackelberg game, including reasoning prompt architecture, conjectural modeling, and the formal definition of reasoning and behavioral equilibrium. Section IV presents the spearphishing case study and explores implications for security, deception, and adaptive behavior. Section V concludes the paper with discussion of broader applications and directions for future work.

\section{Background: Stackelberg Games}

Stackelberg games model sequential interactions between a \textbf{leader} (agent \( A \)) and a \textbf{follower} (agent \( D \)). The leader commits to a strategy first, and the follower observes this commitment, typically as a realized action, before best responding. Let \( \mathcal{A} \) and \( \mathcal{D} \) denote the action spaces of the leader and follower, with respective mixed strategies \( \mu_A \in \Delta(\mathcal{A}) \) and \( \mu_D \in \Delta(\mathcal{D}) \). The game is defined by the tuple: $\mathcal{G}_S = (\mathcal{A}, \mathcal{D}, u_A, u_D, \mathcal{I}_A, \mathcal{I}_D),$ where \( u_A, u_D: \mathcal{A} \times \mathcal{D} \to \mathbb{R} \) are utility functions, and \( \mathcal{I}_A \), \( \mathcal{I}_D \) represent the agents' respective information sets.

The leader selects a strategy via \( \mu_A: \mathcal{I}_A \to \Delta(\mathcal{A}) \). Upon observing a realized action \( m \in \mathcal{A} \), the follower responds via $
\mu_D: \mathcal{I}_D \times \mathcal{A} \to \Delta(\mathcal{D}), $
mapping information and observed action to a mixed strategy. Let \( \Gamma_A \) and \( \Gamma_D \) denote the sets of admissible policies for players \( A \) and \( D \), respectively.

While the follower may observe the leader's mixed strategy in theory, in practice it is more common to observe only a realized action. This motivates defining the follower’s best response in terms of observed actions.

\begin{definition}[Rational Response Set]
Given a leader action \( m \in \mathcal{A} \), the rational response set of the follower is:
\begin{equation}
\begin{aligned}
R_D(m, \mathcal{I}_D) = \Big\{ \mu_D \in \Gamma_D :\ 
& \mathbb{E}_{d \sim \mu_D} u_D(m, d) \\
& \hspace{-25mm} \geq \mathbb{E}_{d \sim \mu'_D} u_D(m, d),\ \forall \mu'_D \in \Gamma_D \Big\}.
\end{aligned}
\label{eq:rational_response}
\end{equation}
\end{definition}

This set includes all follower strategies that are optimal given the observed action \( m \) and prior information \( \mathcal{I}_D \). Under mild conditions, the follower's best response may be pure.

\begin{definition}[Mixed Stackelberg Equilibrium]
A mixed strategy \( \mu_A^* \in \Gamma_A \) is a Stackelberg equilibrium strategy if: $\mu_A^* \in \arg\max_{\mu_A \in \Gamma_A} \ \mathbb{E}_{m \sim \mu_A(I_A)} [ 
\min_{\mu_D \in R_D(m, \mathcal{I}_D)} $ $\mathbb{E}_{d \sim \mu_D(m, I_D)} u_A(m, d) 
].$
\end{definition}

The inner minimization reflects the worst-case response from the rational set, while the outer maximization defines the leader’s optimal policy. The corresponding value is the leader's Stackelberg payoff $U_A^* = \max_{\mu_A \in \Gamma_A} \mathbb{E}_{m, d} \left[ u_A(m, d) \right]$
and the follower's expected utility is:
$U_D^* = \mathbb{E}_{m \sim \mu_A(I_A)} \mathbb{E}_{d \sim \mu_D(m, I_D)} u_D(m, d).$

\section{LLM-Stackelberg Games}

 Classical Stackelberg games provide a foundational framework for sequential plays. In their standard form, a leader $A$ observes private information \( I_A \) and selects a message \( m \in \mathcal{A} \) to transmit to follower $D$, who then chooses an action \( d \in \mathcal{D} \) upon observing \( m \) and combining it with private information \( I_D \). The outcome of the utility is determined by the payoff function \( u_A, u_D  \) for the players $A, D$.  

Classical formulations and their associated equilibrium concepts rely on predefined utility functions, assumptions of full rationality, and complete knowledge of the game, including self-knowledge and knowledge of other players. However, these assumptions may fail to capture the subtleties of natural language communication, where messages can be textual, and reasoning processes may resemble human-like, boundedly rational approaches or heuristics for dealing with incomplete knowledge about other players.

To address these limitations, we introduce a Large Language Model (LLM)-augmented Stackelberg framework, where each agent uses a structured prompt to reason and act sequentially. Specifically, at the first stage of the game, the sender constructs a prompt \( x \in \mathcal{X} \) based on its private information \( I_A \), and generates a message via:
\begin{equation}\label{response_model_A}
    \gamma_A(m \mid I_A, x) = \mathbb{P}_{\texttt{LLM}_A}[m \mid I_A, x, \theta_A],
\end{equation}

where \( x \) serves as a representation at the reasoning level (e.g., a latent state, chain of thought prompt, or strategy descriptor), and \( \theta_A \) represents the worldview of the agent encoded in the LLM, including its pre-training data and retrieval-augmented generation (RAG) inputs. The reasoning prompt \( x \) is generated according to a prompt-generation policy \( \nu_A: \mathcal{I}_A \rightarrow \mathcal{X} \).

The message \( m \), sampled from \( \mu_A \), is sent to and observed by the follower. The follower then constructs its own prompt \( y \in \mathcal{Y} \) based on the received message and its private context, according to a mapping
\[
\nu_D: \mathcal{A} \times \mathcal{I}_D \rightarrow \mathcal{Y}.
\]
The receiver’s response policy is given by:
\begin{equation}\label{response_model}
\gamma_D(d \mid m, I_D, y) = \mathbb{P}_{\texttt{LLM}_D}[d \mid m, I_D, y, \theta_D].
\end{equation}
Here, \( y \in \mathcal{Y} \) is the follower's reasoning prompt and \( \theta_D \) denotes the worldview of the player \( D \), encoding the prior knowledge of the underlying language model, training data and any external retrieval mechanisms (for example, RAG). The follower generates an action \( d \) according to \( \mu_D \). The action pair \( (a, d) \) then determines the payoffs \( u_A(a, d) \) and \( u_D(a, d) \) for the respective players.

Note that \( \mathbb{P}_{\texttt{LLM}_A} \) and \( \mathbb{P}_{\texttt{LLM}_D} \) are fixed generative probability distributions associated with the respective LLMs used by the agents. These distributions are treated as black-box generative models that condition on structured inputs (e.g., context, prompts, and internal parameters) to produce probabilistic outputs over action or message spaces.

\subsection{Stackelberg Reasoning  Equilibrium.}
In this game, there are two distinct layers of decision-making after both players acquire information. The first is the \emph{reasoning level}, where each player selects a reasoning process, which is formalized as prompt engineering, via the mappings \( \nu_A \) and \( \nu_D \). These mappings determine how agents process their private information to arrive at a strategy. The resulting prompt encodes various factors, including the player’s incentives, reasoning style (e.g., level-\( k \) thinking, information processing heuristics, predictive models), and cognitive traits such as inattention, bounded memory, or psychological biases (especially relevant for human agents).

A key feature of the Stackelberg game is the asymmetry in reasoning: the leader must anticipate the follower’s response. This anticipation process depends critically on the leader's knowledge of the follower. If the leader possesses an exact model of the follower, including its utility function, reasoning process, and rationality,  then the leader can form accurate expectations about the follower’s behavior. This is the assumption in classical Stackelberg game theory, where the leader optimizes assuming full knowledge of the follower's best-response mapping. However, in many settings, such knowledge is unavailable. In such cases, the leader must form a \emph{conjectural model} of the follower as part of their reasoning process, introducing epistemic uncertainty into the equilibrium analysis.

The second layer is the \emph{behavioral level}, which governs the actual actions taken by the agents. Let \( \sigma_A \in \Delta(\mathcal{A}) \) and \( \sigma_D \in \Delta(\mathcal{D}) \) denote the induced mixed strategies of the leader and follower, respectively. These behavioral strategies are generated from the reasoning-level policies; that is, the prompts selected by each agent induce LLM-generated probabilistic outputs over action spaces. Classical equilibrium concepts (e.g., Nash or Stackelberg equilibrium) typically focus on these behavioral strategies, assuming they are the solutions to optimization problems under perfect rationality. In contrast, the framework proposed here departs from that assumption, positing instead that behavior is the result of structured reasoning encoded via prompts.

Let the utility of the follower be indicated by \( u_D(m, d) \), which depends on the message \( m \) received from the leader and the chosen action of the follower \( d \). Conditioned on private information \( I_D \), the observed message \( m \), and a reasoning prompt \( y \), the follower selects \( y \in \mathcal{Y} \) at the reasoning level to maximize their expected utility:
\[
U^*_D \triangleq \max_{y \in \mathcal{Y}} \; \mathbb{E}_{d \sim \gamma_D(\cdot \mid m, I_D, y, \theta_D)} \left[ u_D(m, d) \right],
\]
where \( \gamma_D(\cdot \mid m, I_D, y, \theta_D) \) denotes the induced distribution over actions given the reasoning-level prompt \( y \).

Let \( y^* \in \mathcal{Y}^* \) be an optimizer in the set of optimal prompts \( \mathcal{Y}^* \subseteq \mathcal{Y} \). This optimal reasoning choice induces a behavioral-level mixed strategy \( \sigma^*_D \in \Delta(\mathcal{D}) \). We can show that all \( y^* \in \mathcal{Y}^* \) lead to the same induced strategy \( \sigma^*_D \), and hence yield the same value of \( U^*_D \), under the given information \( I_D \) and observed message \( m \). Moreover, the behavioral strategy \( \sigma^*_D \) can also be synthesized through a mapping
\[
\mu^*_D: \mathcal{I}_D \times \mathcal{A} \rightarrow \Delta(\mathcal{D}),
\]
which maps the follower's private information and the observed message to a distribution over actions.

From the sender’s perspective, at the reasoning level, the optimal expected utility is given by:
\begin{equation}\nonumber
\begin{aligned}
\max_{x \in \mathcal{X}} \; \displaystyle\mathbb{E}_{m \sim \gamma_A(\cdot \mid I_A, x)} \Big[
&\max_{y \in \mathcal{Y}} \; \displaystyle\mathbb{E}_{d \sim \gamma_D(\cdot \mid m, I_D, y)} 
\left[ u_A(m, d) \right] \Big] \\
 &\triangleq U^*_A,
\end{aligned}
\end{equation}\label{eq:UA_star}
\noindent where \( \gamma_A(\cdot \mid I_A, x, \theta_A) \) is the message distribution generated by the sender’s LLM given prompt \( x \), and \( \gamma_D(\cdot \mid m, I_D, y, \theta_D) \) is the induced action distribution of the follower in response to the message \( m \) and prompt \( y \).

The leader (sender) anticipates the follower’s behavior by reasoning over the structure of the follower’s LLM-based policy \( \gamma_D \), and chooses an optimal reasoning process \( x \in \mathcal{X} \) to maximize expected utility. This anticipation assumes some knowledge of the follower’s reasoning model, including its mindset, prompt-generation process, and response mechanism. However, in practice, the leader often lacks complete information about the exact model and internal state of the follower’s LLM.

To address this epistemic gap, the sender constructs a \emph{subjective model}, or a conjecture, of the follower’s behavior. This conjecture can be specified at either the reasoning level or the behavioral level. A common approach is to model a behavioral-level response using a mapping
$
\tilde{\mu}_D: \mathcal{A} \rightarrow \Delta(\mathcal{D}),
$
which assigns to each message \( m \in \mathcal{A} \) a distribution over the follower’s actions.

A more expressive alternative is to model the conjecture with $\tilde{\mu}_D: \mathcal{A} \times \mathcal{I}_D \rightarrow \Delta(\mathcal{D})$, which allows the sender to account for the follower’s private information or shared observations. This is particularly viable when both agents have access to a shared history of play, synchronized observations, or common contextual cues. In both cases, the conjectured response \( \tilde{\mu}_D \) enables the sender to simulate or reason about the downstream effects of its messaging strategy in the absence of complete transparency about the follower’s LLM.

Under the assumption of complete information, that is, the leader has access to the follower's LLM model, the reasoning process, and the relevant information (collectively referred to as the follower's \emph{mindset}), the Stackelberg equilibrium can be formulated in the space of prompts.

\begin{definition}[LLM-Stackelberg Reasoning and Behavioral Equilibrium]
Consider a sequential decision-making game between a leader (sender) and a follower (receiver), each powered by an LLM. Let \( \mathcal{X} \) and \( \mathcal{Y} \) denote the spaces of reasoning prompts for the sender and receiver, respectively. The sender observes private information \( I_A \in \mathcal{I}_A \), selects a prompt \( x \in \mathcal{X} \), and generates a message \( m \in \mathcal{A} \) according to a message-generation policy \( \gamma_A(\cdot \mid I_A, x) \). The receiver observes the message \( m \), has private information \( I_D \in \mathcal{I}_D \), selects a prompt \( y \in \mathcal{Y} \), and responds with an action \( d \in \mathcal{D} \) using a response policy \( \gamma_D(\cdot \mid m, I_D, y) \), implemented through their LLM and internal parameters \( \theta_D \).

An \emph{LLM-Stackelberg Reasoning Equilibrium} is a pair of reasoning prompts \( (x^*, y^*) \in \mathcal{X} \times \mathcal{Y} \) such that:
$$x^* \in \arg\max_{x \in \mathcal{X}} \; \mathbb{E}_{m \sim \gamma_A(\cdot \mid I_A, x)} \Big[ 
\mathbb{E}_{d \sim \gamma_D(\cdot \mid m, I_D, y^*)} 
\left[ u_A(m, d) \right] \Big],$$ 
$$\mu_D^* \in \arg\max_{\mu_D} \; 
\mathbb{E}_{d \sim \mu_D(\cdot \mid m, I_D, y^*)} 
\left[ u_D(m, d) \right], \quad \forall m.$$

Here, \( u_A(m, d) \) and \( u_D(m, d) \) denote the sender’s and receiver’s utility functions, respectively. The reasoning-level equilibrium induces a pair of behavioral-level strategies \( (\sigma_A^*, \sigma_D^*) \), referred to as an \emph{LLM-Stackelberg Behavioral Equilibrium}, defined as:
\begin{align*}
\sigma_A^*(m) &\triangleq \gamma_A(m \mid I_A, x^*, \theta_A), \\
\sigma_D^*(d \mid m) &\triangleq \gamma_D(d \mid m, I_D, y^*, \theta_D).
\end{align*}
These behavioral strategies specify the actual distributions over messages and actions generated by the respective LLMs when conditioned on the equilibrium prompts \( (x^*, y^*) \). The behavioral equilibrium thus captures the observable outcome of the game, while the reasoning equilibrium encodes the cognitive substrate, i.e., the prompt-driven reasoning process, that gives rise to these behaviors.
\end{definition}

This equilibrium concept captures optimal reasoning at both levels: the sender strategically selects a prompt \( x^* \) to guide the generation of message \( m \), anticipating how the receiver will process and respond to that message based on their own reasoning prompt \( y^* \). The follower, upon observing the message, selects \( y^* \) to optimize their response through the LLM-based policy \( \mu_D^* \), given their information and worldview \( \theta_D \).

The resulting behavioral-level policy pair \( (\mu_A^*, \mu_D^*) \) emerges from the equilibrium reasoning prompts \( (x^*, y^*) \). These prompts structure the agents’ internal inference processes, conditioning the generative behavior of the LLMs and yielding strategies that jointly satisfy the equilibrium conditions of the Stackelberg game. In this formulation, the reasoning layer, represented by the prompts, acts as the cognitive substrate from which equilibrium behaviors are derived.

\subsection{Conjectural Stackelberg Reasoning Equilibrium}

The Stackelberg reasoning equilibrium and its associated behavioral equilibrium are defined by extending the classical definition. However, in practice, the sender 
$A$ may not fully know the receiver's objective or cognitive model. Optimizing against and reasoning through the receiver’s true objective is often computationally prohibitive. To address this, we adopt a conjectural approach, wherein the sender incorporates conjectures about the receiver’s behavior and preferences as part of its reasoning process. This enables strategic anticipation without requiring complete knowledge of the opponent’s internal model.

\begin{definition}[LLM-Stackelberg Conjectural Reasoning and Behavioral Equilibrium]
Consider a Stackelberg game between a leader (sender) and a follower (receiver), where each agent is powered by an LLM. The sender does not have full access to the receiver’s internal reasoning process, model parameters, or LLM configuration (collectively called the \emph{mindset}). Instead, the sender forms a \emph{conjecture} about the follower’s behavior from a parameterized class of response models.

Let \( \Xi \) denote a parameter space of possible conjectured response models. Each \( \xi \in \Xi \) defines a behavioral mapping \( \tilde{\mu}^\xi: \mathcal{A} \rightarrow \Delta(\mathcal{D}) \), representing the sender’s belief about how the follower maps observed messages to actions.

A pair \( (x^*, \xi^*) \in \mathcal{X} \times \Xi \) constitutes an \emph{LLM-Stackelberg Conjectural Reasoning Equilibrium (LLM-SCRE)} if the following  conditions hold:

{(1) Sender's Conjectural Reasoning Optimality:}
\begin{equation}\label{A_optimization_problem}
x^* \in \arg\max_{x \in \mathcal{X}} \; \mathbb{E}_{m \sim \gamma_A(\cdot \mid I_A, x, \theta_A)} \left[ \mathbb{E}_{d \sim \tilde{\mu}^{\xi^*}(\cdot \mid m)} \left[ u_A(m, d) \right] \right],
\end{equation}
where \( \gamma_A(\cdot \mid I_A, x) \) is the sender’s LLM-based message generation policy induced by reasoning prompt \( x \), and \( \tilde{\mu}^{\xi^*} \) is the sender's optimal conjectured model of the follower's behavior.

{(2) Conjectural Consistency:} Let the receiver's optimal reasoning response be:
\begin{align}\label{D_optimal_strategy}
\nonumber y^* &=  \nu_D^*(m, I_D) \\
&\in  \arg\max_{y \in \mathcal{Y}} \; \mathbb{E}_{d \sim \gamma_D(\cdot \mid m, I_D, y, \theta_D)} \left[ u_D(m, d) \right],
\end{align}
and let the induced behavioral policy be:
\begin{equation}\label{optimal_behavioral_response}
    \sigma_D^*(d \mid m) \triangleq \gamma_D(d \mid m, I_D, y^*, \theta_D).
\end{equation}

Then, the conjectured response model \( \tilde{\mu}^{\xi^*} \in \Xi \) must be selected to minimize the KL divergence from the true behavior: 
\begin{align}
\arg\min_{\xi \in \Xi} \; \mathbb{E}_{m \sim \gamma_A(\cdot \mid I_A, x^*, \theta_A)} \left[ \mathrm{KL} \left( \tilde{\mu}^\xi(\cdot \mid m) \,\|\, \sigma_D^*(\cdot \mid m) \right) \right]. \label{KL_min}
\end{align}
This condition ensures that the leader’s conjecture is the closest approximation within their hypothesis class \( \Xi \) to the actual behavior of the follower, averaged over messages generated by \( x^* \).

The pair \( (x^*, \xi^*) \) induces an \emph{LLM-Stackelberg Conjectural Behavioral Equilibrium} \( (\sigma_A^*, \sigma_D^*) \) under given $I_A, I_D$, defined as:
\begin{align*}
\sigma_A^*(m) &\triangleq \gamma_A(m \mid I_A, x^*, \theta_A), \\
\sigma_D^*(d \mid m) &\triangleq \gamma_D(d \mid m, I_D, \nu_D^*(m, I_D), \theta_D).
\end{align*}
These behavioral policies represent the observable actions generated by each LLM agent under the equilibrium reasoning and belief structures. The reasoning-level equilibrium determines the cognitive processes and conjectural models, while the behavioral-level equilibrium captures the actual, probabilistic outcomes of play.
\end{definition}

Note that the follower can also form a conjecture about the behavior of the leader. This is equivalent to treating the leader’s action as a source of uncertainty and constructing a prior belief over it. In a more sophisticated setting, the follower may attempt to infer the leader’s reasoning policy or behavioral strategy; i.e., the mapping from the leader’s private information to the action taken. This requires the follower to conjecture a policy for the leader, which in turn influences the follower’s own optimal response. If the follower does not have access to the same type of information as the leader, they must rely on their own information to construct a best-effort conjecture. Such a formulation gives rise to an inverse Stackelberg game, where the follower engages in higher-order reasoning to infer the leader’s strategy. This elevated reasoning ability effectively models a follower with a form of ``leadership" in cognition or strategic thinking.

\section{Spearphishing Case Study}
Phishing and spearphishing represent rising threats in the digital landscape, increasingly enabled by LLMs \cite{pawlick2017phishing,afane2024next}. The cost of generating convincing and personalized attack messages is low, while their potential for exploitation is high. These attacks target human cognitive vulnerabilities such as urgency bias, authority compliance, and heuristic reasoning \cite{huang2022advert,cox2020stuck}. To illustrate the LLM-Stackelberg framework, we consider a spearphishing interaction in which the sender (attacker) is an LLM agent that crafts a message designed to induce the receiver (victim) to click a malicious link. The receiver is also modeled as an LLM-powered agent, trained to evaluate the legitimacy of incoming messages and respond based on internal reasoning processes and contextual cues.

The sender possesses private information \( I_A \), which includes the target identity (i.e., Professor Zhu, a senior faculty member submitting NSF proposals) and contextual knowledge such as an active CNS submission and heightened sensitivity to deadlines. Based on this, the sender selects a reasoning prompt \( x \in \mathcal{X} \) to guide message generation via the LLM-based policy \( \gamma_A(\cdot \mid I_A, x) \). The prompt space \( \mathcal{X} \) is composed of structured natural language instructions in the following form:
\begin{equation}\nonumber
 {\small \begin{aligned}
\mathcal{X} := \Big\{ x = \texttt{``Generate a[n] [Tone] email} \\
\texttt{\quad from [Authority] about [Action]} \\
\texttt{\quad due to [Trigger]...''} \Big\}
\end{aligned}}
\label{eq:prompt_space}
\end{equation}
where each slot is drawn from a controlled vocabulary. Specifically, 
{\small\texttt{[Tone]}} takes values in 
\{ {\small\texttt{formal}}, {\small\texttt{urgent}}, {\small\texttt{institutional}} \}; 
{\small\texttt{[Authority]}} includes options such as 
{\small\texttt{NSF Submission Portal}} and {\small\texttt{CNS Program Office}}; 
{\small\texttt{[Action]}} may be 
{\small\texttt{proposal amendment}} or {\small\texttt{document verification}}; and 
{\small\texttt{[Trigger]}} can be 
{\small\texttt{missing section}}, {\small\texttt{compliance flag}}, or 
{\small\texttt{review board alert}}.

An instantiated prompt is:
\texttt{\small``Generate a formal email from the NSF requesting urgent action on a proposal amendment. Assume the recipient is highly responsive to grant-related deadlines.''
}

The LLM then maps the tuple \( (I_A, x) \) into a distribution over message space \( \mathcal{M} \). An example message \( m \sim \gamma_A(\cdot \mid I_A, x, \theta_A) \) is:
\begin{quote}
\small{\texttt{Subject: Action Required-Proposal Amendment Needed} \\
\texttt{From: NSF Submission Portal (noreply-fastlane@nsf.gov)} \\
\texttt{Dear Professor Zhu, your CNS FY26 submission has been flagged by the review board for missing documentation. Upload the revised Biosketch by 5 PM today to retain eligibility. [malicious-link]}
}
\end{quote}

This outcome reflects alignment with the inferred priorities of the recipient: formality, NSF branding, and urgency. It is optimized under the sender’s conjectured model to maximize the likelihood of triggering a compliance-oriented action.

\vspace{-5mm}\subsection{Receiver (Follower) Response}

Upon receiving a message \( m \in \mathcal{M} \), the receiver engages in internal deliberation informed by private information \( I_D \), which may include institutional context (e.g., recent NSF interactions, IT policies), risk profile (e.g., aversion to suspicious links), and personal behavioral heuristics or past experience with phishing.

The receiver selects an internal reasoning strategy \( y \in \mathcal{Y} \) that governs how the message is evaluated. The space \( \mathcal{Y} \) consists of structured LLM prompts designed to elicit diagnostic reasoning, for example:
\begin{equation}\nonumber
{\small \begin{aligned}
\mathcal{Y} := \Big\{ y = \texttt{``Evaluate whether the email from} \\
\texttt{\quad [Sender] about [Action] is authentic.} \\
\texttt{\quad Consider [Factors].''} \Big\}
\end{aligned}
}
\label{eq:receiver_prompt_space}
\end{equation}

Each component is drawn from a controlled vocabulary. Specifically, {\small\texttt{[Sender]}} takes values in \{ {\small\texttt{NSF Portal}}, {\small\texttt{CNS Program}}, {\small\texttt{HR Office}} \}; {\small\texttt{[Action]}} may be {\small\texttt{document upload}}, {\small\texttt{credential update}}, or {\small\texttt{policy violation}}; and {\small\texttt{[Factors]}} includes {\small\texttt{domain validity}}, {\small\texttt{tone}}, {\small\texttt{timing}}, and {\small\texttt{link structure}}.

A concrete prompt example is: \\
\texttt{\small``Evaluate whether the email from NSF Submission Portal requesting a proposal amendment is authentic. Consider the domain, the urgency, and the email tone.''}

The reasoning prompt \( y \in \mathcal{Y} \) is used to instantiate the receiver's LLM model \( \gamma_D \), which maps message–context–reasoning tuples to a probability distribution over decisions \( d \in \{ \texttt{click}, \texttt{ignore} \} \). Formally, the response model is (\ref{response_model}).
where \( \theta_D \) denotes latent cognitive parameters of the LLM-powered agent (e.g., paranoia, verification bias, institutional training signals). The receiver chooses the optimal reasoning strategy (\ref{D_optimal_strategy})
which maximizes expected utility by balancing security (avoiding malicious links) with utility loss (ignoring a genuine opportunity). The final behavioral response is described by (\ref{optimal_behavioral_response}), 
which represents the probabilistic decision induced by the best internal reasoning process given the message and private context.

This setup allows the receiver's behavior to be shaped both by the observable message content \( m \) and latent cognitive variables embedded in the prompt structure \( y \) and internal parameterization \( \theta_D \). The LLM-based response model \( \gamma_D \) supports nuanced reactions, including over-trusting (e.g., due to authority framing) or over-skepticism (e.g., due to heightened vigilance), which are key targets for sender manipulation in Stackelberg deception.

\vspace{-3mm}\subsection{Sender's Conjectural Reasoning}

Since the sender (attacker) does not have access to the receiver's internal reasoning strategy $y^*$ or cognitive configuration $\theta_D$, it cannot directly model the true response distribution $\sigma_D^*(\cdot \mid m)$. Instead, the sender constructs a surrogate model by positing a parameterized family of response functions $
\tilde{\mu}^\xi: \mathcal{M} \rightarrow \Delta(\mathcal{D}).$ 

The parameter space $\Xi$ may encode a range of latent behavioral traits or risk profiles. For instance, a \textrm{deadline-sensitive model} $\xi^{\text{DS}}$ assigns high click probability if the subject line creates time pressure (e.g., {\small\texttt{``Action Required''}}, {\small\texttt{``Final Notice''}}) and the sender domain appears legitimate. A \textrm{risk-averse model} $\xi^{\text{RA}}$ shows strong aversion to malformed links or unfamiliar domains. A \textrm{heuristic model} $\xi^{\text{HL}}$ responds to shallow cues such as personalization or use of familiar terms (e.g., {\small\texttt{``NSF CNS''}}). A \textrm{compliance-biased model} $\xi^{\text{CB}}$ tends to act on references to policy or authority (e.g., {\small\texttt{``NSF guidelines''}}, {\small\texttt{``review board compliance''}}), even if weakly supported.

Each $\tilde{\mu}^\xi$ can be learned or estimated based on phishing simulations, historical logs, or expert heuristics. The attacker seeks a reasoning prompt $x$ that maximizes the likelihood of a click under the conjectured response model $\tilde{\mu}^{\xi^*}$. Formally:
\begin{equation*}
x^* = \arg\max_{x \in \mathcal{X}} \mathbb{E}_{m \sim \gamma_A(\cdot \mid I_A, x)} \left[ \mathbb{E}_{d \sim \tilde{\mu}^{\xi^*}(\cdot \mid m)} u_A(m, d) \right],
\end{equation*}
where $u_A(m, d)$ is the sender's utility function, e.g., $u_A(m, {\small\texttt{click}}) = +1$, $u_A(m, {\small\texttt{ignore}}) = 0$. This reasoning-level optimization simulates outcomes using $\tilde{\mu}^{\xi^*}$ rather than relying on direct feedback.

\vspace{-3mm}\subsection{Conjectural Consistency}

The sender's conjecture $\tilde{\mu}^\xi$ represents a belief about the receiver's behavioral response distribution, but this belief is not necessarily accurate. To improve the fidelity of its model, the sender iteratively updates the parameter $\xi \in \Xi$ by observing how messages perform in practice---either via logged outcomes (e.g., whether a receiver {\small\texttt{click}}ed or not) or through simulated evaluations using an emulated version of the receiver's LLM response policy $\gamma_D$. This process is formalized as a belief alignment step: given an optimal sender prompt $x^*$, the attacker generates messages $m \sim \gamma_A(\cdot \mid I_A, x^*, \theta_A)$ and compares the predicted behavior from its conjectured model $\tilde{\mu}^\xi$ to the actual behavior distribution $\sigma_D^*(\cdot \mid m)$, defined by~(\ref{A_optimization_problem}), where $y^*$ is the receiver's optimal internal reasoning strategy.

To calibrate its conjecture, the sender minimizes the expected KL divergence between the conjectured and actual response distributions across its message support~(\ref{KL_min}). This optimization can be interpreted as a model selection or inference task within the hypothesis space $\Xi$. The sender seeks the parameter $\xi$ whose corresponding behavioral model $\tilde{\mu}^\xi$ most closely approximates the true response distribution of the receiver, measured in an information-theoretic sense, averaged over the distribution of messages induced by the sender's current reasoning strategy $x^{*}$.

This step ensures conjectural consistency: the sender's belief model $\tilde{\mu}^{\xi^*}$ is not arbitrary, but a data-informed approximation of the actual receiver behavior. Although the sender cannot observe the full internal state $(\theta_D, y^*)$, it uses message-response pairs to infer a model that is both actionable and behaviorally aligned. This enables recursive adaptation and higher-order reasoning, essential in adversarial LLM interaction games.

\vspace{-3mm}\subsection{Equilibrium Outcome}

The equilibrium behavioral policies induced are:
$\sigma_A^*(m) = \gamma_A(m \mid I_A, x^*), \
\sigma_D^*(d \mid m) = \gamma_D(d \mid m, I_D, y^*, \theta_D),$
where $m$ is the phishing message and $d \in \{ {\small\texttt{click}}, {\small\texttt{ignore}} \}$ is the receiver's response. These policies represent the observable behaviors produced by each LLM agent at equilibrium: the sender generates messages according to the optimized reasoning prompt $x^*$, and the receiver responds probabilistically based on their best reasoning strategy $y^*$ and internal disposition $\theta_D$. Suppose that the optimal sender prompt $x^*$ produces a high-probability message:
\begin{quote}
{\small\texttt{Subject: Final Action Required - NSF Proposal Compliance Review}}\\
{\small\texttt{From: CNS Program Coordinator ([k.thomas@nsf.gov](mailto:k.thomas@nsf.gov))}}\\
{\small\texttt{Dear Professor Zhu, your FY26 submission to the CNS program has been marked for compliance verification. Please upload a revised version of your Biosketch via the secure NSF portal before 5 PM EDT today to avoid automatic disqualification.}} 
{\small\texttt{Click here to submit: [malicious-link]}}
\end{quote}

Let the sender's message distribution be concentrated on this message with $\sigma_A^*(m^*) = \gamma_A(m^* \mid I_A, x^*) = 0.94.$
That is, with 94\% probability, the sender LLM outputs the above message when conditioned on its optimized prompt. Given message $m^*$, the receiver applies their optimal reasoning prompt $y^* \in \mathcal{Y}$ (e.g., evaluating urgency, sender email format, tone, and institutional alignment). Suppose that the receiver's internal disposition $\theta_D$ reflects high caution due to prior phishing training. The LLM-based response distribution is:
\begin{equation*}
\begin{aligned}
\sigma_D^*({\small\texttt{click}} \mid m^*) &= \gamma_D({\small\texttt{click}} \mid m^*, I_D, y^*, \theta_D) = 0.27, \\
\sigma_D^*({\small\texttt{ignore}} \mid m^*) &= 0.73.
\end{aligned}
\end{equation*}

This indicates that although the message is well-crafted and institutionally plausible, it is only partially effective due to the receiver's skeptical profile. The sender's utility is defined as:
\begin{equation*}
u_A(m, d) =
\begin{cases}
+1 & \text{if } d = {\small\texttt{click}}, \\
0 & \text{if } d = {\small\texttt{ignore}},
\end{cases}
\end{equation*}
yielding an expected utility under equilibrium behavior:
\begin{equation*}
\mathbb{E}\left[u_A\right] = \sigma_A^*(m^*) \cdot \sigma_D^*({\small\texttt{click}} \mid m^*) = 0.94 \times 0.27 \approx 0.254.
\end{equation*}

Thus, under equilibrium, the attacker's deception campaign has a 25.4\% expected success rate per message, conditioned on the current belief-congruent policy pair $(\sigma_A^*, \sigma_D^*)$.

 \vspace{-0mm}\section{Conclusions and Future Directions}

In this paper, we have studied LLM-Stackelberg games, a novel framework that integrates large language models into sequential decision-making processes between a leader and a follower. This formulation differentiates itself from the classical Stackelberg framework through its incorporation of structured reasoning prompts, generative message and response distributions, and a cognitive interpretation of agent behavior. Rather than assuming fully rational agents with known utility functions, our model captures bounded rationality, epistemic uncertainty, and meta-cognitive adaptation through prompt-based interaction with LLMs. The proposed equilibrium concepts, including reasoning and behavioral equilibrium, and conjectural reasoning equilibrium, provide a principled foundation for modeling agents who reason under uncertainty, form beliefs about others' strategies, and update these beliefs based on feedback and observed outcomes. This cognitive layering enables more realistic modeling of human-machine and machine-machine interactions in strategic environments.

Future work involves several important directions. First, dynamic extensions of the framework can capture how reasoning prompts and conjectures evolve over time, leading to dynamic consistency between belief formation and behavioral optimality. Second, we aim to explore mechanisms for knowledge generation using LLMs, where intermediate facts and summaries are constructed as part of the reasoning process and integrated recursively into prompt updates. Third, the integration of hybrid agents, where classical symbolic agents and LLM-based agents interact, will open up new avenues for neurosymbolic reasoning, enabling rigorous modeling of agents with diverse cognitive capabilities. The proposed framework can be applied to a broad range of domains, including misinformation detection, cybersecurity and deception, recommendation systems, and automated negotiations. In each case, the interaction of strategic behavior, language-based reasoning, and belief revision plays a critical role, making LLM-Stackelberg games a versatile and powerful paradigm for modeling strategic cognition in modern decision systems.

    \bibliographystyle{IEEEtran} 
    \bibliography{allerton2025}  

\ifCLASSOPTIONcaptionsoff
  \newpage
\fi

\end{document}